\documentclass[conference]{IEEEtran}
\IEEEoverridecommandlockouts
\usepackage{cite}
\usepackage{amsmath,amssymb,amsfonts}
\usepackage{algorithmic}
\usepackage{graphicx}
\usepackage{textcomp}
\usepackage{tabularx}
\usepackage{booktabs}
\usepackage{enumitem}
\usepackage{listings}
\usepackage{xcolor}
\definecolor{light-gray}{gray}{0.95}
\usepackage[
 format=plain, 
 justification=justified, 
 labelfont=small, 
 textfont=small 
 ]{caption}
\renewcommand{\thetable}{\Roman{table}}
 
\def\BibTeX{{\rm B\kern-.05em{\sc i\kern-.025em b}\kern-.08em
    T\kern-.1667em\lower.7ex\hbox{E}\kern-.125emX}}

\begin{document}

\title{Integration of Domain Expert-Centric Ontology Design into the CRISP-DM for Cyber-Physical Production Systems\\
\vspace{-0,2cm}
}
\author{\IEEEauthorblockN{Milapji Singh Gill, Tom Westermann, Marvin Schieseck, Alexander Fay}
\IEEEauthorblockA{\textit{Institute of
Automation Technology},\\
\textit{Helmut-Schmidt-University}\\
Hamburg, Germany \\
{\{milapji.gill, tom.westermann, marvin.schieseck, alexander.fay\}@hsu-hh.de}}
\vspace{-1cm}
}\maketitle

\begin{abstract}
    
In the age of Industry 4.0 and Cyber-Physical Production Systems (CPPSs) vast amounts of potentially valuable data are being generated. 
Methods from Machine Learning (ML) and Data Mining (DM) have proven to be promising in extracting complex and hidden patterns from the data collected. The knowledge obtained can in turn be used to improve tasks like diagnostics or maintenance planning. 
However, such data-driven projects, usually performed with the Cross-Industry Standard Process for Data Mining (CRISP-DM), often fail due to the disproportionate amount of time needed for understanding and preparing the data. 
The application of domain-specific ontologies has demonstrated its advantageousness in a wide variety of Industry 4.0 application scenarios regarding the aforementioned challenges. However, workflows and artifacts from ontology design for CPPSs have not yet been systematically integrated into the CRISP-DM. Accordingly, this contribution intends to present an integrated approach so that data scientists are able to more quickly and reliably gain insights into the CPPS. The result is exemplarily applied to an anomaly detection use case.\\ \end{abstract}

\newcommand{\uproman}[1]{\uppercase\expandafter{\romannumeral#1}}

\section{Introduction}\label{introduction}
The constantly increasing complexity and varying demands as a result of rapidly changing customer needs require flexible production systems \cite{E.Jarvenpaa.2016}. 
The transformation to a CPPS through systematic digitalisation and automation has the potential to support companies concerned on their path towards an "intelligent manufacturing" in order to cope with frequently changing requirements \cite{Dogan.2021, Hildebrandt.2020}. 
With the help of CPPSs processes can be accelerated and costs reduced \cite{Hildebrandt.2018}. In this context, the efficient management of data, information and knowledge is an essential cornerstone for a successfully operating production system \cite{VogelHeuser.2021, Dogan.2021,VDIVDE.2022}. Consequently, the amount of data that accumulates in a CPPS must be collected and structured systematically. In this way, new knowledge can be discovered, stored in a knowledge base and used for a wide variety of application scenarios \cite{Dou.2015, Zhang.2018, VogelHeuser.2021}. DM or ML methods offer the potential to extract implicit knowledge from the data intending to incorporate it for decision support \cite{UsamaFayyad.1996, Dogan.2021, VDIVDE.2022}. 

However, the actual implementation of such data-driven projects entails a vast number of obstacles in the industrial environment. Particularly data scientists are confronted with a multitude of problems when attempting to utilise the data in a value-adding manner \cite{Yin.2015}. These arise in various phases of such projects which often follow the CRISP-DM, a quasi-standard process model \cite{Wirth., Huber.2019}. Especially the phase of data understanding and data preparation take up a disproportionate amount of the total project time \cite{Dou.2015, P.C.Soto.2021b}. This is due, among other things, to the handling and integration of diverse heterogeneous data sources in Information and Operational Technology (IT \& OT) which can differ in their syntactic, structural and semantic characteristics \cite{Hildebrandt.2020, Dogan.2021, VogelHeuser.2021}. Further challenges are e.g. missing metadata, inconsistent data or incomplete information \cite{VogelHeuser.2021, Dou.2015}. While present in other domains as well, these problems are especially prevalent in CPPS where many different systems, each using their own data models and interfaces, need to interact \cite{Jirkovsky.2017, Gill.2022}. Moreover, the lack of domain knowledge of data scientists to interpret differing relationships (e.g. the physical interactions for diagnostics) delays the progress \cite{Zhang.2018}. Data-driven projects are based on cross-disciplinary cooperation between various domain experts. Thus, the effective inclusion and exchange of knowledge are crucial for success. However, currently no standard defines how this knowledge can be formalized, documented and shared between experts in the CRISP-DM. With standardized approaches for preparing and preprocessing the data on grounds of domain knowledge, it can be assumed that this time effort could be reduced significantly \cite{MartinezPlumed.2017, Saltz.1215202112182021}. For all these problems, the use of domain-specific ontologies from the field of symbolic AI with defined concepts and relations can offer considerable added value \cite{Jirkovsky.2017, Zhang.2018, Dogan.2021}. Hence, domain-specific ontological artifacts (OAs), created by different experts during the ontology design for CPPS, need to be integrated efficiently into the CRISP-DM. Efficient in this regard means accelerating the steps of data understanding and data preparation in the long term by providing modular, reusable and semantically unambiguous OAs. 

Accordingly, the following research question is to be answered in this contribution: 
\textit{"How can the expert-centric design of ontologies with its created ontological artifacts be integrated into the CRISP-DM workflow so that data scientists are able to more quickly and reliably gain insights in CPPS?"}

To address this research question, the paper is structured as follows: Sec. \ref{background} first introduces the CRISP-DM as well as the domain expert-centric ontology design for CPPS. Subsequently, requirements regarding the enhancement of the CRISP-DM are derived and used to analyze related works in Sec. \ref{requirements and related works}. The main contribution of this paper, the extended CRISP-DM, is presented in \ref{contribution}. This combined approach is applied in Sec.\ref{use case} in the course of a use case for anomaly detection in a hybrid mixing plant. In the concluding Sec. \ref{summary} the results are summarised and an outlook is given.

\section{Background}\label{background}

\subsection{CRISP-DM}\label{CRISP-DM}
Discovering knowledge in databases is a non-trivial process of identifying valid, novel, potentially useful and understandable patterns in data \cite{UsamaFayyad.1996}. The modeling itself is just one step in the process of discovering knowledge in databases. Moreover, it is a creative approach requiring several different steps and skills from distinct experts. The cross-disciplinary cooperation of various experts such as data scientists, process or control engineers is a premise to conduct ML or DM projects in CPPS successfully. Thus, different methodologies including a defined sequence of steps were introduced to provide a more holistic view of the knowledge discovery process\cite{Huber.2019, Wirth.}. The CRISP-DM is a generic and widely adopted de-facto standard. This process model describes a framework for translating business problems into DM tasks and carrying out data-driven projects independent of both the domain of interest as well as the technology used \cite{Huber.2019, VDIVDE.2022, MartinezPlumed.2017, Schroer.2021, Wirth.}. Basically, the CRISP-DM consists of six successively performed steps. However, individual steps can be repeated depending on the circumstances and any new evidence so that the desired result is attained. The first step is the business understanding during which the project goals are determined. In the second step, the data understanding, hypotheses for hidden information with respect to the DM project goals are formulated based on the experience of domain experts alongside assumptions. Accordingly, the relevant data gets collected and prepared in the third step. The modeling takes place in the fourth step on the basis of the prepared and preprocessed data. Hereafter, the identified patterns or the trained model are evaluated in step five. Besides, the results are assessed according to the underlying business objectives. After the successful evaluation deployment into the business environment takes place in the last step \cite{Wirth.}.  

Despite the widespread use of CRISP-DM, there are a number of deficiencies that need to be addressed with regard to its application in CPPS. In particular, the steps of data understanding and data preparation are complex and time-consuming. In this respect, the integration of domain-specific knowledge from experts is especially advantageous as a means to accelerate the comprehension, integration, selection and preparation of data \cite{VogelHeuser.2021, P.C.Soto.2021b, Zhou.2021}. This domain-specific knowledge can be provisioned in the form of ontologies \cite{Dou.2015, Hildebrandt.2018, Zhang.2018}.
\subsection{Domain-Expert-Centric-Approach to Ontology Design}\label{Ontology Design}
 With regard to the efficient development and the flexible application of domain-specific ontologies in CPPS for different data-driven projects, some requirements need to be considered. First of all, a modular and extensible ontology is essential as the design can be very complex and time-consuming\cite{Hildebrandt.2018}. In this respect, the reusability of created OAs (e.g. lightweight ontology design pattern (ODP), T-Box, A-Box) in different projects is crucial.  Consequently, there is a need for a defined procedure with regard to designing as well as extending existing ontologies for the specific information needs of a data-driven project. Primarily, this can reduce the effort to model individual ontologies for a specific project. With the intention of providing a semantically unambiguous definition of individual concepts and their relations, domain-specific industry standards should be used for modeling ontologies in CPPS. Besides, it is vital that the ontology design follows a domain expert-centric approach. Domain expert-centric in this context means that the respective domain experts take on a developing role in the modeling of pertinent concepts and relations. In this way, the knowledge can be fully formalised and integrated into the CRISP-DM. 

The approach of Hildebrandt et al. \cite{Hildebrandt.2018, Hildebrandt.2020} proves to be particularly beneficial in CPPS with regard to these requirements. 
In this context, the development of a formal ontology for CPPS in a machine-interpretable format (e.g. ontology web language (OWL)), also called heavyweight ontology (HWOs), is characterised by three methodological building blocks (MBBs). Additionally, individual input and output artifacts as well as roles (domain expert, ontology expert and end user) are depicted. 
In the first MBB the general project requirements including user stories, collected in a purely informal artifact (e.g. text files or tables), are needed to initiate the ontology design. On this basis so-called competency questions (CQs) with further relevant context information (e.g. data sources) are recorded in an ontology requirements specification document by the domain expert and the end user. These serve as the output of the first MBB. With the aim of describing the expert knowledge in a reusable way, relevant information resources of the domain (e.g. domain-specific industry standards) and pre-existing lightweight ODPs (in the form of UML class diagrams) are required in the second MBB. Again, the end user and the domain expert are in a developing role. The result is an aligned lightweight ontology (LWO), formed to satisfy the requirements specified before. This, together with existing heavyweight ODPs, forms the basis for the third MBB in which the HWO is built. The previously elaborated OAs serve as a template for the ontology expert in order to develop and implement the necessary concepts and relations in OWL. In this course, the T-Box is modeled and mappings are defined for the population of the A-Box. The ODPs newly generated can in turn be reused which allows to built individual ontologies faster in subsequent projects. 
For the reasons mentioned, this method is suitable for the integration into the CRISP-DM workflow in the following.

\section{Requirements and Related Work}\label{requirements and related works}
\subsection{Requirements}\label{requirements}
With regard to the systematic incorporation of domain knowledge about CPPS into the CRISP-DM, essential requirements have to be considered:

\textbf{R1 (Seamless integration of domain expert-centric ontology design steps)}: The consideration of domain-specific ontologies in data-driven projects for data understanding and preparation is helpful \cite{Dou.2015, Zhang.2018, VogelHeuser.2021}. Typically, each project requisites a specific ontology that has to be modeled, adapted or extended individually. Thus, the steps or MBBs from the domain expert-centric ontology design with created and reusable OAs must be integrated and documented sensibly and seamlessly into the existing workflow of the CRISP-DM. In this way, project-specific ontologies can be built efficiently. 

\textbf{R2 (Provision of understandable, modular and reusable OAs)}: The OAs developed during the method are intended to support data scientists in carrying out their tasks. Thus, it is substantial to provide understandable as well as interpretable OAs (e.g. by means of  UML class diagrams) in order to be used beneficially afterwards \cite{Dou.2015, Hildebrandt.2018}. For this purpose, it is advisable to model OAs based on industry standards (with concepts and relations known by the community) which are of particular importance in the course of data understanding and preparation. Moreover, the OAs are provided in different degrees of formality depending on the desired goal (e.g. partial automation of individual steps). In the course of this, an OA flow is to be added and integrated to the CRISP-DM. Additionally, with the aim of reducing the effort of creating project-specific ontologies in the long term, the OAs should be modular and reusable \cite{Hildebrandt.2020}.  

\textbf{R3 (Definition of experts and roles)}: For the successful implementation of a data-driven project, the interaction of various experts in different roles is necessitated\cite{MartinezPlumed.2017, Saltz.1215202112182021, Hildebrandt.2020}. Accordingly, a list of relevant experts must first be determined who possess know-how about the CPPS, the necessary technologies as well as methods from sub-symbolic and symbolic AI. Additionally, it is imperative to include the end users as a means to capture the information needs. Since the first step is not yet about deployment, required roles for implementing software (e.g. software developers) are initially neglected. Throughout the course of the project, the experts are assigned to individual method steps and take on either a developing or a consulting role. A consulting role means that the experts are rather available in an advisory capacity. A developing role, by contrast, implies that the experts are directly involved in the creation of OAs. 






\subsection{Related Work}\label{relatedWorks}

In the following, related work will be analysed. 
While there are various contributions examining a domain-unspecific approach to integrate ontologies in data-driven projects (e.g. \cite{XinWang.2005}), this paper mainly focuses on the domain of Cyber-Physical Systems (CPS) or CPPS. Hence, publications considering the integration of domain-specific ontologies in the steps up to and including data preparation are contemplated. 


In the analysis, it is noticeable that a variety of publications contain isolated approaches to integrating domain-specific ontologies into DM or ML (e.g. \cite{Westermann.2022, Zhang.2018}). Vogel-Heuser et al. \cite{VogelHeuser.2021} e.g. outline the potentials of combining semantics with data analysis in the context of digital twins. In this course, challenges in data analysis and the creation and use of ontologies are delineated first. Subsequently, applications are depicted in which the combination of both fields offers potential. A feasibility check is carried out using the example of a file system crawler and forum mining for ontology extension. Nevertheless, a systematic method that combines the two approaches including necessary steps, reusable OAs as well as required roles is not explicitly specified.

An ontology-based approach for preprocessing data and enhancing data quality for ML use cases is proposed by \cite{P.C.Soto.2021b}. In this course, an ontology is used in both the data understanding as well as in the data preparation. In the data understanding the ontology is consulted to understand the data required for the ML use case in the semiconductor industry. For this purpose, a packing ontology with subclasses and individuals is automatically generated. In addition, data cleaning steps are performed using reasoning and SPARQL queries. In comparison to this contribution, no modular,  industry standards-based and reusable OAs (e.g. ODPs) are integrated into the process. Beyond that, there is no designation of required roles.

Zhou et al. \cite{Zhou.2021} introduce SemML, a software system that allows to reuse and generalise ML pipelines for condition monitoring of production processes. This is realized by relying on ontologies and ontology patterns. The tool chain presented aims to facilitate the laborious tasks of ontology creation, data preparation and ML modeling across stakeholders with differing backgrounds. Furthermore, the developed ontologies are used as a suitable means for communication between experts of different backgrounds. Looking at the OAs used, they are neither modular nor based on industry standards. As outlined before, the use of industry standards-based ODPs promises a more efficient reusability in a multitude of data-driven projects. In addition, this contribution aspires to be applicable to a variety of different use cases that go beyond condition monitoring.  

In summary, there is no approach for CPPS that considers the methodological steps of both disciplines coherently. This contribution aims to close this research gap.

\section{Contribution} \label{contribution}

Intending to answer the initial research question posed, the following Sec. \ref{contribution} presents a workflow that combines both the CRISP-DM with the domain expert-centric approach to ontology design. 
The steps of modeling, evaluation and deployment are omitted firstly. 

\subsubsection*{Methodological Overview} 
\begin{figure*} [h]
    \centering
    \includegraphics[width=0.8\textwidth]{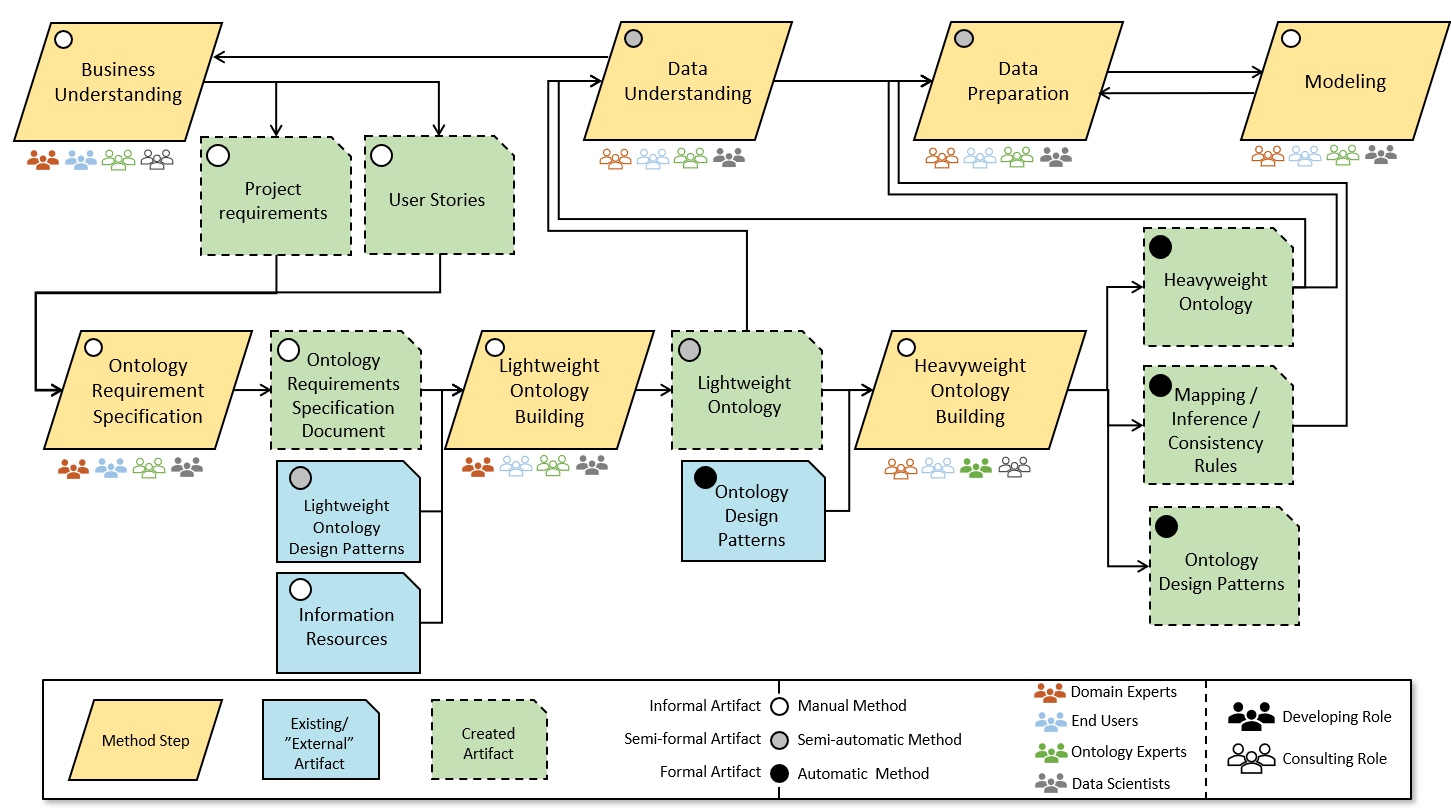}
    \caption{Integration of domain-expert-centric ontology design into the CRISP-DM}
    \label{fig:OntoDM-Workflow}
\end{figure*}

The complete workflow depicts all the steps already included in the regular CRISP-DM along with three additional method steps from the domain expert-centric approach to ontology design (see \textbf{R1}). These have been incorporated between the steps of business understanding and data understanding (see Fig. \ref{fig:OntoDM-Workflow}). The integration at this point lends itself to various reasons. Firstly, it allows to consider the project requirements and the business problem in the ontology requirements specification step. Secondly, it ensures that the generated OAs are available to the data scientists in the subsequent steps. In principle, activities from data understanding and preparation are transferred to the domain expert-centric ontology design. The advantage is the long-term reusability of the generated OAs. Moreover, the sequence of steps follows the one defined in the original CRISP-DM. Therefore, regressions are possible, e.g. after the phase of data understanding. The proposed workflow involves four different developing or consulting roles with specific expertise (see \textbf{R3}). The end user is involved to describe the information needs and elicit user stories. Due to the heterogeneous nature and the many interacting systems in the domain of CPPS the necessary knowledge to satisfy these information needs is usually spread out across multiple experts. Hence, the role of the domain expert is mandatory who is familiar with the domain of the CPPS and the system under consideration. With the aim of designing the HWO requested, an ontology expert needs to be included. Finally, and in deviation from the roles defined in \cite{Hildebrandt.2020}, the data scientist, possessing knowledge about DM and ML, is needed. 
In this approach, the created OAs (e.g. LWO and HWO) are intended to support the data scientist in the data understanding and data preparation steps in order to reduce particular manual and time-consuming activities. The reusability of industry standards based ODPs in future projects is potentially helpful in reducing the extensive manual workload (e.g. initial creation of the ontology) in the long-term by partially automating manual activities. This especially applies to the process of data cleansing and preparation. In this regard, the application of semantic rules to the ontology can help to automatically identify and correct erroneous or incomplete data as well as checking consistency. Basically, the OAs are classified according to their degree of formality. Besides, a further subdivision of the OAs takes place. On the one hand, to those that were created during the specific project (project-specific OAs in green). On the other hand, to those that have already originated in previous data-driven projects (in blue, see \textbf{R2}). The integrated workflow including all OAs is elucidated separately in detail below.

\subsubsection*{Business Understanding}
During the business understanding step, domain experts in collaboration with end users define the objectives of the data-driven project. Data scientists and ontology experts can provide consulting support. Domain experts are particularly suitable for this task not least because of the extensive understanding of the business objective as well as the possible impact of the data-driven project. In addition to classic tasks in this step (business problem analysis, project requirements etc.) user stories are set up by the end users to determine the information needs in a standardised way. The user stories should cover information on requirements, pertinent stakeholders, the purpose of the ontology, relevant knowledge, data sources as well as acceptance criteria. 
If multiple user stories are defined for the same use case, the resulting ontology should ideally be able to cover all of them. 
\subsubsection*{Ontology Requirement Specification}
During the ontology requirement specification, the end users, the domain experts and the data scientists translate the user stories into CQs. These are the essential parts of the ontology requirements specification document and are intended to cover all information demanded. 
In this context, thoughts should be made about a pre-selection of relevant features for the modeling step. Therefore, domain experts and data scientists can build up on their knowledge of possible relationships in the CPPS under consideration as well as hypothetical models. 
 If the ontology is to be used for automated activities of preprocessing, e.g. by integrating inference mechanisms or rules, this should also be specified.
\subsubsection*{Lightweight Ontology Design}
In the next step a LWO is to be built based on the ontology requirements specification document. The domain experts and the data scientists are in a developing role. This is due to the extensive knowledge about appropriate concepts of the domain as well as the necessary standards. The procedure here can be divided into two parts. Firstly, it is advisable to check whether there are existing lightweight ODPs (e.g. from previous projects) modeled as UML class diagrams. These lightweight ODPs include all concepts and relations that are needed to answer the aforementioned CQs. Secondly, in the case of missing model elements, suitable information resources must be consulted to model new lightweight ODPs with UML class diagrams. Again, to increase reusability, references need to be made to information resources like industry standards and norms. In this context, one lightweight ODP should be based on one standard. Subsequently, the individual lightweight ODPs are to be aligned into a project-specific LWO. The four mechanisms \emph{equivalent-to, sub-classing, relation-to, attribute-to-class} are suitable to accomplish this endeavour \cite{Hildebrandt.2018}. 
\subsubsection*{Heavyweight Ontology Design}
During the HWO design, the T- and A-Box are modeled by ontology experts. The LWO from the previous step provides all the necessary concepts and relations for this task. The data scientists, the domain experts and the end users provide advisory support rather than being in a developing role. The ontology experts first check whether existing ODPs created in previous projects can be reused. In the absence of such OAs, the modeled LWO is used as a basis for developing the HWO. Similarly, in this case, an ODP should be assigned to one industry standard. The alignment of individual ODPs is achieved through the use of the mechanisms presented before. The T-Box modeling is performed with suitable tools (e.g. Protege), followed by the population of the A-Box with instances. The implementation can be executed manually or with mapping languages such as RML or R2RML. The decision here depends on the data format provided. If, for example, recorded measured values are required the data can also be virtualised instead of being materialised. SPARQL queries can be used to ascertain whether the CQs are answered sufficiently. The HWO is the essential OA that emerges during this step. Beyond that, depending on the purpose of the ontology, it is appropriate to provide further OAs. These can be, for example, the scripts for executing the mappings. If rules or constraints have been modeled in SHACL or SWRL, these should also be saved and reused.
\subsubsection*{Data Understanding}
In the data understanding phase, data scientists can consult the created OAs. In particular, the designed LWO in the form of the UML class diagrams is an OA that can be interpreted comfortably by data scientists. This OA is intended to help the comprehension of the modeled knowledge as well as the interrelated information in the HWO. Furthermore, the HWO as well as SPARQL queries are available with easier access to the data about the CPPS. On this basis, data scientists can perform exploratory analysis and derive conclusions for further activities in the CRISP-DM. Additionally, scripts with defined mappings as well as consistency rules, that have been used for data integration, are also made available as formal OAs. With regard to reusability, for each new data-driven project existing OAs should be checked. In the long term, this can reduce the time required for data access. 
\subsubsection*{Data Preparation} Also during data preparation data scientists, who are still in the developing role, benefit from the preliminary work. Since the data sources have already been merged meaningfully in the HWO, the effort of data integration is significantly reduced. Hereby, subsequent activities such as data cleaning and feature engineering can be performed. By using SPARQL queries, the required data records can be generated quickly and with minimal effort. If not already done during ontology design, pertinent features for the subsequent modeling can be determined. The ontology inherently focuses on the extensibility of data. In case further data sources need to be included, they can be added to the graph without changing the existing structure. Besides, ontologies offer the possibility of using logic reasoners to check the present graph triples for logic consistency. While the reasoners can't change detected inconsistencies, they can recognize possible errors that were committed during the ontology design or the mappings. In the best case, the integrated rules as well as consistency checks also reduce the effort required to clean the data.

\section{Exemplary Use Case} \label{use case}
\subsection{Use Case: Hybrid Mixing Process}

\begin{figure}[htbp]
\centering
\includegraphics[width=0.4\textwidth]{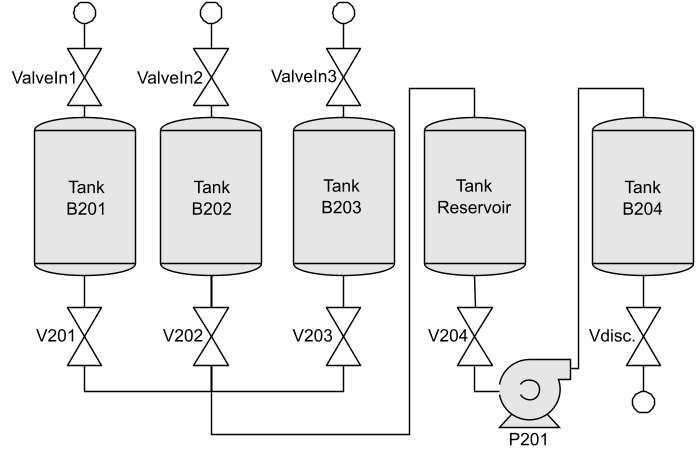}
\caption{Three tank mixing module}
\label{fig:MixingModule}
\end{figure}

The integrated approach, outlined in Sec \ref{contribution}, has been applied to an anomaly detection use case in a mixing process (see Fig. \ref{fig:MixingModule}). This process continually mixes three kinds of liquids from separate tanks into a reservoir. It is controlled by numerous actuators that can open or close valves as well as turn on the pump between the tanks. Multiple sensors record changes in variables like filling level or flow through the pipes. Both the states of the actuators (e.g. open/close) and any values recorded by the sensors are stored inside a relational database. Additional engineering information is present in various engineering documents (e.g. P\&I diagrams). 
\vspace{-0,3cm}

\paragraph*{Business Understanding}
An unplanned breakdown of the mixing module results in a variety of costs. Early identification of anomalous behaviour, e.g. caused by clogged pipes, is potentially helpful to improve maintenance planning. Hence, during the business understanding step the plant engineer (domain expert) defined project requirements, as depicted in Sec. \ref{contribution}. The end user, in this case a maintenance technician, described the information needs in the form of user stories that guided the following ontology design steps. One exemplary user story raised in this context was: \textit{'As a maintenance technician, I would like to analyze the systems timing behaviour, so that the detection of anomalous behaviour is possible'}


\paragraph*{Ontology Requirement Specification}
Based on the user stories, the plant engineer, the maintenance technician as well as the data scientist formulated CQs that the HWO was supposed to answer. An excerpt of these questions can be seen in Table \ref{tab:CQ:HybridMixingProcess}.

\begin{figure} [ht]
    \centering
    \includegraphics[width=0.4\textwidth]{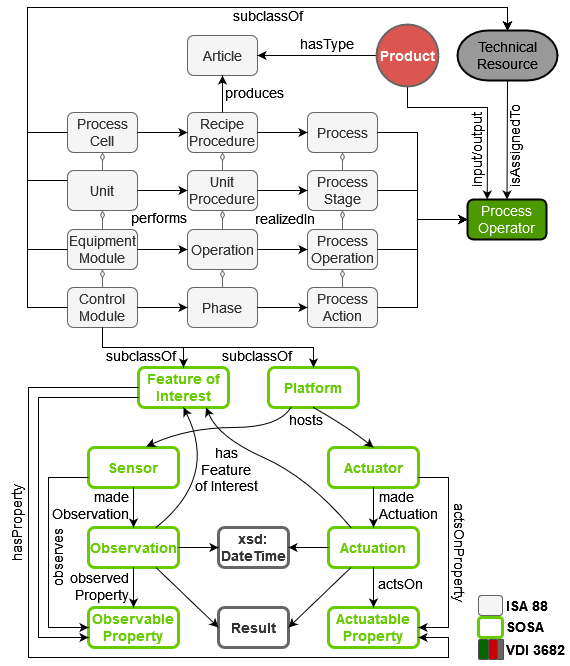}
    \caption{LWO of the hybrid mixing process}
    \label{fig:LightWeightOntologyOntoDMME}
    \vspace{-0,5cm}
\end{figure}

\begin{table}[ht]
  \centering
  \begin{tabularx}{0.45\textwidth}{X|p{0.15\textwidth}|p{0.05\textwidth}}
    \textbf{Competency Question} & \textbf{Answer} & \textbf{ODP} \\
    \toprule
    Which sensors are part of Tank B201? & tank\_B201.level, B201\_isFull & ISA88, SOSA\\
    Which part of the plant performs the process of filling Tank B201? & FillEmptyTankB201 & VDI3682, ISA88 \\
    ... & ... & ... \\
  \end{tabularx}
\caption{Excerpt of CQs from the use case}
\label{tab:CQ:HybridMixingProcess}
\end{table}
\vspace{-0,3cm}
\paragraph*{Lightweight Ontology Building}
Using the set of CQs as a guideline, the plant engineer and the data scientist identified the set of lightweight ODPs that reflected the necessary concepts for the LWO. A distinction is made between CQs that could be answered directly based on the data source and those that had to be determined by the data scientist using DM and ML methods.
Accordingly, it was essential to provide information regarding the hybrid mixing process available in a structured and uniform manner for the data understanding. Since all ODPs are either based on industrial standards (e.g. VDI 3682, ISA 88) or are widely accepted by the community (SOSA - e.g. description of sensors and observations), the resulting ontology offers high reusability. It is worth mentioning that the ODPs VDI 3628 (formalized process description), ISA 88 (plant hierarchy and  process recipes) and DIN EN 61360 (properties) have already been created in past projects\cite{Hildebrandt.2020, Westermann.2022}. In this respect, the effort required to create the LWO was negligible (see Fig. \ref{fig:LightWeightOntologyOntoDMME}).

\paragraph*{Heavyweight Ontology Building}
During HWO building, the LWO as well as pre-existing ODPs were accessed by the ontology expert in order to model the T-Box. For the hybrid mixing plant, the ODPs were aligned according to the LWO and populated in two different ways: First, the static knowledge from engineering documents was mapped into the ontology using  the mapping language \emph{RML}\cite{Dimou.2014} and the mapping interpreter \emph{SDM-RDFizer} \cite{Iglesias.2020}. 
Additionally, the observations and actuations, namely the actual timestamps and measurements from sensors and actuators, needed to be mapped into the ontology. As stated earlier, these measurements were stored in a relational database. Consequently, an ontology-based data access using \emph{Ontop}\cite{Calvanese.2016} was chosen and mappings between the relational database and the ontology were defined. 
This approach has several benefits. For example, it keeps the persistently stored knowledge graph compact and avoids duplication of data.

\paragraph*{Data Understanding}
During the data understanding step, the data scientist used the LWO and the HWO from previous steps to lighten his workload. 
The LWO from Fig. \ref{fig:LightWeightOntologyOntoDMME} offered an easily accessible description of the domain-specific concepts as well as their relations which helped in the first understanding of the data and its structure. It also aided in finding existing relationships in the data for further analysis by modeling relevant prior knowledge. In this use case, the LWO showed that process cells can perform predefined processes which are structured hierarchically. Moreover, it revealed that control modules could host various sensors and actuators that in turn observe properties of some feature of interest. In contrast to this, the HWO alongside the implemented ontology-based data access 
could be utilized during data exploration. Using SPARQL, a highly flexible access to the measurements was possible i.e. to all sensors of a technical resource that performed a certain process or all sensors that measured variables related to tank B201 (s. Listing \ref{list:ObservationsB201}). Since the selection criteria originally stem from different data silos (e.g. sensor measurements in a relational database and engineering knowledge from their respective documents) exploration like this would only have been possible after a large data integration effort which was now shifted to the design phase of the ontology. In addition to the data access, the ontology also offered a straightforward way of documenting the data using comments and other annotations. By involving established vocabularies like ECLASS \cite{Eclass.14.04.2023} the properties of a system could be unambiguously described facilitating the model transfer between different stakeholders. 

\begin{lstlisting}[basicstyle=\footnotesize, breaklines=true, backgroundcolor=\color{light-gray}, captionpos=b, caption=SPARQL query selecting all measurements from sensors that have Tank B201 as their feature of interest, label=list:ObservationsB201,
   basicstyle=\ttfamily]
select ?time ?result ?sensor ?property ?foi
where {
 ?sensor    sosa:observes ?property;
            sosa:madeObservation ?obs.
 ?obs       sosa:resultTime ?time;
            sosa:hasSimpleResult ?result;
            sosa:hasFeatureOfInterest ?foi.
 ?property  ssn:isPropertyOf ?foi. 
 VALUES (?foi) {(ModVA:mixer_partial0.tank_B201)}
 }
\end{lstlisting}

\paragraph*{Data Preparation}
In terms of data integration most work has been done during the design phase of the HWO. Instead of manually joining large amounts of data, the integration was done by the ontology expert using established concepts from industrial standards. At modeling time, all required data could be loaded from the knowledge graph using the ontology-based data access defined before. The open source reasoner Pellet was applied to infer additional formal knowledge about the system based on the explicitly one provided before. 

    

\begin{figure}[ht]
\centering
\includegraphics[width=0.48\textwidth]{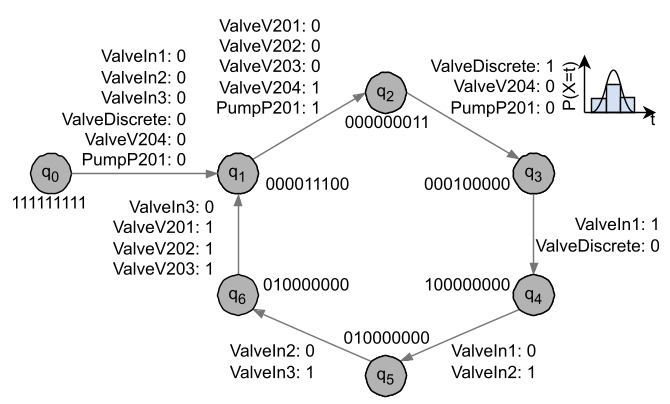}
\caption{Learned automaton of the three tank mixing system showing state vectors, transition events and an exemplary timing distribution}
\label{fig:LearnedTimedAutomaton}
\end{figure}

\paragraph*{Modeling}
During the modeling step, a two-phase approach to model learning was chosen in order to realise anomaly detection. Since the actuations of certain variables (i.e. opening of a valve, or turning on a pump) correspond to changes in the continuous behaviour of the system \cite{O.Niggemann.2015}, this behaviour should be learned separately. Therefore, in the first step, the values of all discrete variables throughout time were needed and could be queried from the HWO. From these variables, a timed automaton was learned using the OTALA-algorithm (see Fig. \ref{fig:LearnedTimedAutomaton}) \cite{A.Maier.2015}. By comparing a previously unseen sequence to the behaviour permitted by the automaton, anomalies in the CPPS' behaviour could be detected with regard to transition sequence and transition time. Using this automaton, it was possible to classify whether any given sequence of events and measurements could be considered. 

\subsection{Artifact reusability in subsequent use cases}
While the modeling effort to design and populate the ontology is still considerable, many of the OAs are reusable in subsequent use cases. In this section, two scenarios are reflected upon (see Table  \ref{tab:ReusabilityOfartifacts}): 
\subsubsection{Scenario A: Different Use Case, Same CPPS}
When investigating an additional use case at the same CPPS, many of the OAs can be reused due to the inherent extensibility of ontologies. 
Assuming that a different use case (e.g. diagnostics) should be investigated, a new user story needs to be defined. Based on this, a new ontology requirement specification is built in which, depending on the user story, some CQs as well as ODPs can be reused. 
During LWO and HWO building, only non-reusable ODPs need to be integrated into the existing model. The same holds for the mapping rules where only new data sources have to be integrated into the HWO. Depending on the number of reusable ODPs and data sources, this can shorten the manual effort in creating the HWO considerably. 

\subsubsection{Scenario B: Same Use Case, different CPPS}
If the same use case should be solved for a different but structurally similar CPPS (i.e. another batch production system), the design process focuses less on extending a prior model and more on mapping new data sources into the previously designed ontology. Most of the OAs can be directly reused from the previous application. 
Like the hybrid mixing process, batch production systems follow the identical hierarchical classification and contain various sensors that record discrete and continuous variables. 
Since it is an equivalent use case, the user story and ontology requirement specifications will be similar. Likewise, it contains the same ODPs, the LWO and alignment ontologies. 
Due to the heterogeneous structure of CPPS data, the mapping rules are usually specific to each system and can therefore not be reused. However, once the data is mapped into the ontology, it can be queried using the same SPARQL queries from the previous use case. 
\begin{table}[ht]
  \centering
  \begin{tabularx}
  {0.49\textwidth}{X|p{0.12\textwidth}|p{0.11\textwidth}}
    \textbf{Ontological Artifact} & \textbf{New Use Case, same CPPS} & \textbf{Same Use Case, Different CPPS} \\
    \toprule
    User Stories & No & Yes \\   
    Ontology Req. Specification & Yes with extension & Yes \\    
    Lightweight Ontology & Yes with extension & Yes\\
    Alignment Ontology & Yes with extension & Yes\\  
    Heavyweight Ontology & Yes with extension & Yes \\    
    Mapping Rules & Yes with extension & No\\        
    SPARQL Queries (Exploration) & Yes with extension & Yes\\
    SPARQL Queries (Preparation) & Yes with extension & Yes\\     
  \end{tabularx}
\caption{Reusability of OAs for additional use cases}
\label{tab:ReusabilityOfartifacts}
\end{table}

\section{Summary and Outlook}\label{summary}
In this contribution an approach was presented for integrating domain expert-centric ontology design into the CRISP-DM for CPPS. The overarching goal was to increase the efficiency of the data scientists' activities in the data understanding and data preparation steps by incorporating formalized domain knowledge. Initially, the two basic methods were introduced to discuss shortcomings of the current version and the advantages of a combination. 
Accordingly, the main contribution delineated an enhanced CRISP-DM with method steps from domain expert-centric ontology design, reusable OAs as well as imperative roles. The integrated approach was demonstrated using an anomaly detection use case involving a hybrid process plant where a timed automaton was learned using the OTALA algorithm.

Concerning the initial research question posed, the following can be stated: The presented contribution has demonstrated, as in various previously published works, that the use of domain-specific ontologies is advantageous in data-driven projects. With regard to efficiency, it was essential that the steps with created and already existing OAs were integrated seamlessly before data understanding and preparation. In reference to the domain expert-centric approach to ontology design, it was also necessary to define roles with specific task profiles throughout the process. These requirements have all been taken into consideration in the extended CRISP-DM. Nevertheless, there is still the need to evaluate the approach in further use cases. This is particularly necessary in order to meet non-functional requirements such as validity, reliability, utility and stringency. In this respect, the proposed process must be evaluated based on several different data-driven projects. However, since the two methods are already widely used separately, it can be hypothesised that this could also be the case for the integrated approach presented. In addition, there are further possibilities for expansion. In future work supplementary steps of the CRISP-DM such as modeling and evaluation with created OAs could be improved. Furthermore, research could be conducted to automate some of the current manual steps.



\section*{Acknowledgment}
This research [project ProMoDi and EKI] is funded by dtec.bw – Digitalization and Technology Research Center of the Bundeswehr. dtec.bw is funded by the European Union – NextGenerationEU.

This work has been partially supported and funded by the German Federal Ministry of Education and Research (BMBF) for the project “Time4CPS - A Software Framework for the Analysis of Timing Behaviour of Production and Logistics Processes” under the contract number 01IS20002.

\bibliographystyle{IEEEtran}

\end{document}